\documentclass[lettersize,journal]{IEEEtran}
\usepackage{amsmath,amsfonts}
\usepackage{algorithmic}
\usepackage{algorithm}
\usepackage{array}
\usepackage[caption=false,font=normalsize,labelfont=sf,textfont=sf]{subfig}
\usepackage{textcomp}
\usepackage{stfloats}
\usepackage{url}
\usepackage{colortbl}
\usepackage{verbatim}
\usepackage{graphicx}
\usepackage{makecell}
\usepackage{cite}
\usepackage{caption}
\usepackage{subfloat}
\usepackage{bbding}
\usepackage{color}
\usepackage{multirow}
\usepackage[percent]{overpic}
\usepackage[pagebackref,breaklinks,colorlinks,urlcolor=black]{hyperref}
\usepackage{booktabs}
\usepackage[linewidth=1pt]{mdframed}
\usepackage{arydshln}
\usepackage{soul}
\usepackage[thicklines]{cancel}

\RequirePackage{xspace}

\makeatletter
\DeclareRobustCommand\onedot{\futurelet\@let@token\@onedot}
\def\@onedot{\ifx\@let@token.\else.\null\fi\xspace}

\usepackage{cite}

\RequirePackage{xspace}
\makeatletter
\DeclareRobustCommand\onedot{\futurelet\@let@token\@onedot}
\def\@onedot{\ifx\@let@token.\else.\null\fi\xspace}

\def\ie{\emph{i.e}\onedot} 
 
\def\etc{\emph{etc}\onedot}

\def\etal{\emph{et al}\onedot}
\makeatother
\newcommand{\stdvu}[1]{\scriptsize{(#1)}}  
\usepackage[capitalize]{cleveref}
\crefname{section}{Sec.}{Secs.}
\Crefname{section}{Section}{Sections}
\Crefname{table}{Table}{Tables}
\crefname{table}{Tab.}{Tabs.}
\Crefname{equation}{Equation}{Equations}
\crefname{equation}{Eqn.}{Eqns.}
\usepackage{booktabs}
\usepackage{caption}
\usepackage{subfloat}
\usepackage{bbding}
\usepackage{color}
\usepackage{multirow}
\usepackage{makecell}
\usepackage[percent]{overpic}
\usepackage{bm}

\newcommand{\mliu}[1]{\textcolor[rgb]{0,0,0}{#1}}
\newcommand{\mat}[1]{\bm{\mathit{#1}}}
\newcommand{\matx}{\mat{x}}

\newcommand{\matm}{\mat{m}}
\newcommand{\matc}{\mat{c}}
\newcommand{\mati}{\mat{i}}
\newcommand{\mata}{\mat{a}}
\newcommand{\mate}{\mat{e}}
\newcommand{\ii}[1]{\mathit{#1}}
\newcommand{\bb}[1]{\mathbf{#1}}
\begin{document}
\captionsetup[figure]{labelfont={bf},labelformat={default},labelsep=period,name={Fig.}}
\title{Retrieval Augmented Image Harmonization}

\author{Haolin Wang, Ming Liu, Zifei Yan, Chao Zhou, Longan Xiao, Wangmeng Zuo,~\IEEEmembership{Senior Member,~IEEE}
	\thanks{Haolin Wang, Ming Liu, Zifei Yan, and Wangmeng Zuo are with the School of Computer Science and Technology, Harbin Institute of Technology, Harbin 150001, China (e-mail: \href{mailto:why_cs@outlook.com}{why\_cs@outlook.com}, \href{mailto:csmliu@outlook.com}{csmliu@outlook.com}, \href{mailto:yanzifei.hit@gmail.com}{yanzifei.hit@gmail.com}, \href{mailto:wmzuo@hit.edu.cn}{wmzuo@hit.edu.cn}).
		
		Chao Zhou and Longan Xiao are with the TRANSSION, Shenzhen 518038, China (e-mail: \href{mailto:chao.zhou5@transsion.com}{chao.zhou5@transsion.com},\href{mailto:longan.xiao1@transsion.com}{longan.xiao1@transsion.com}).}}

\markboth{Journal of \LaTeX\ Class Files,~Vol.~14, No.~8, August~2021}%
{Shell \MakeLowercase{\textit{et al.}}: A Sample Article Using IEEEtran.cls for IEEE Journals}

\IEEEpubid{0000--0000/00\$00.00~\copyright~2021 IEEE}

\maketitle

\begin{abstract}
When embedding objects (foreground) into images (background), considering the influence of photography conditions like illumination, it is usually necessary to perform image harmonization to make the foreground object coordinate with the background image in terms of brightness, color, and \etc.
Although existing image harmonization methods have made continuous efforts toward visually pleasing results, they are still plagued by two main issues.
Firstly, the image harmonization becomes highly ill-posed when there are no contents similar to the foreground object in the background, making the harmonization results unreliable.
Secondly, even when similar contents are available, the harmonization process is often interfered with by irrelevant areas, mainly attributed to an insufficient understanding of image contents and inaccurate attention.
As a remedy, we present a retrieval-augmented image harmonization (Raiha) framework, which seeks proper reference images to reduce the ill-posedness and restricts the attention to better utilize the useful information.
Specifically, an efficient retrieval method is designed to find reference images that contain similar objects as the foreground while the illumination is consistent with the background. 
For training the Raiha framework to effectively utilize the reference information, a data augmentation strategy is delicately designed by leveraging existing non-reference image harmonization datasets.
Besides, the image content priors are introduced to ensure reasonable attention.
With the presented Raiha framework, the image harmonization performance is greatly boosted under both non-reference and retrieval-augmented settings.
The source code and pre-trained models will be publicly available.

\end{abstract}

%

\section{Introduction}

One common process of image composite is integrating an object (denoted as foreground) from an image into another one (denoted as background). Due to different photography conditions like illumination, inharmoniousness inevitably exists between the foreground and background in terms of brightness, color, and \etc.
As a remedy, image harmonization aims to adjust the appearance of the foreground object and make it visually consistent with the background image. 
Traditional methods typically extract hand-crafted statistics to adjust the foreground appearance, which often fails in complex scenarios. With the advances of deep learning, a series of DNN-based methods are proposed to learn the translation in a data-driven manner. For example, \cite{DoveNet,Bargainnet,RAIN} predict dense-to-dense translations for image harmonization. Some methods also explore high-resolution image harmonization by learning comprehensible image filters~\cite{DCCF,Harmonizer}, piecewise curve mapping~\cite{liang2021spatial}, and RGB-to-RGB transformation~\cite{CDTNet}.

\begin{figure}[t]
    \centering
    \includegraphics[width=\linewidth]{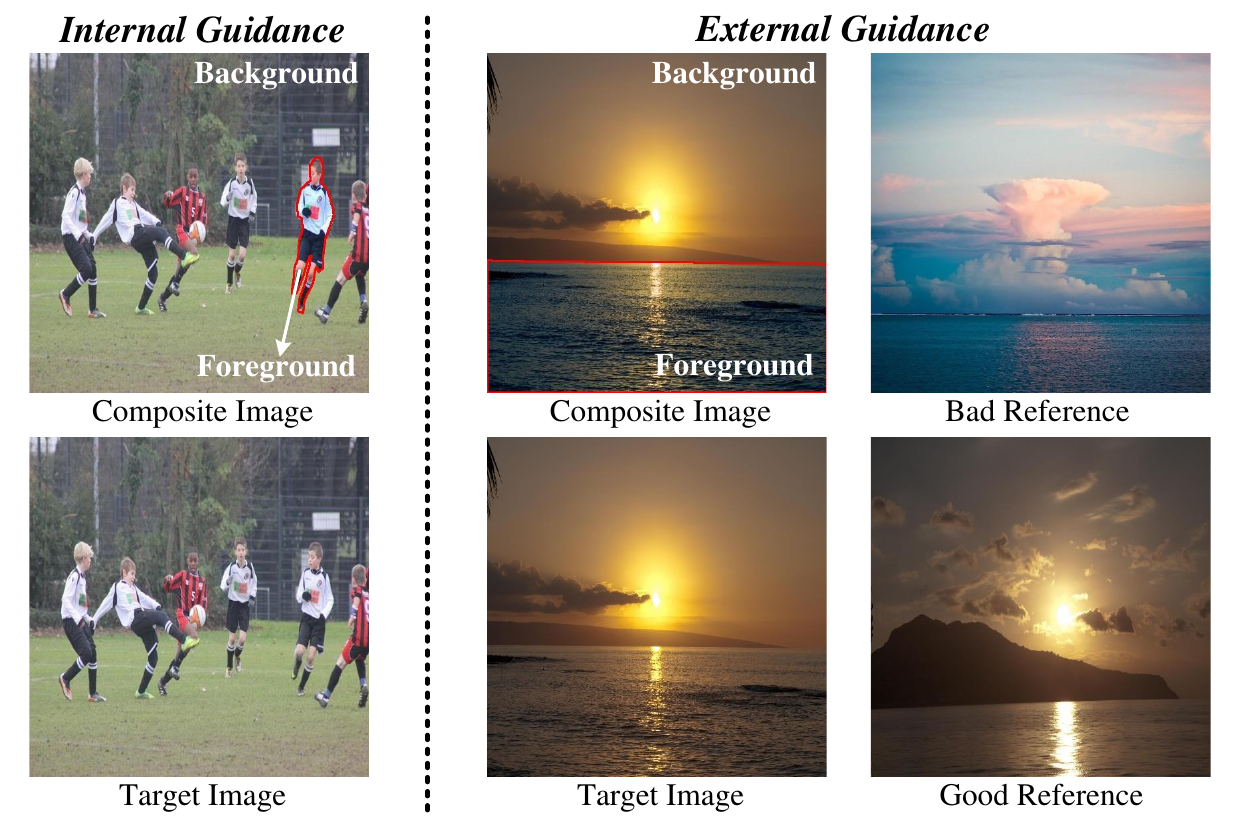}
    \caption{When the background contains similar content to the foreground, it can provide internal guidance for adjusting foreground appearance. Conversely, image harmonization becomes ill-posed. Providing references that have similar content to the foreground can alleviate this issue. However, while the illumination between the reference and background is different, it fails to provide reliable guidance. Thus, when a reference has a consistent illumination condition with the background, the appearance of the foreground-similar content can provide reliable guidance, as it has a similar appearance to the target image.}
    \label{fig:intro}
\end{figure}

Recently, the community has noticed that objects in the background image that are similar to the foreground can provide guidance for image harmonization (see \cref{fig:intro}).
For example, Zhu~\etal~\cite{MRR} propose a Localtion-to-Localtion module to fuse foreground location with the related background feature and a Patches-to-Location Module to further adjust foreground feature at a high-resolution level. BAIN~\cite{Scs-co} is also designed to obtain attention-weighted background feature distribution according to the similarity between foreground and background features. HDNet~\cite{HDNet} introduces a Local Dynamic Module to find $K$ local representations from background regions according to the foreground feature.
However, in the real scenarios of image harmonization, the background image cannot always contain similar content related to the foreground (see \cref{fig:intro}). In this case, image harmonization degrades to the highly ill-posed setting and cannot achieve visually realistic results.

\IEEEpubidadjcol

To compensate for the performance loss caused by the absence of similar objects in the background regions, an intuitive approach is to prepare some reference images through image retrieval methods.
With suitable references and accurate attention, the image harmonization models can accordingly adjust the foreground appearance.
It seems that an appropriate combination of image retrieval methods and cross-attention mechanisms can provide a competent retrieval-augmented image harmonization solution.
Nevertheless, existing image retrieval methods~\cite{chen2022deep,off-the-shelf} ignore the requirements of the image harmonization task.
In such a condition, the differences, especially in terms of illumination, between the background and the retrieved images are often non-negligible.
Therefore, \emph{the first task of this work is to develop an image retrieval method}, where the retrieved reference images should possess two characteristics, \ie,
(i)~there should be objects similar to the foreground in the reference image, and
(ii)~the reference image should share similar illumination to the background image.

%
%

Furthermore, even though proper reference images are available, it is still challenging to utilize the guidance for image harmonization. Existing methods usually deploy cross attention modules~\cite{zhangcontrolvideo} to adjust the foreground appearance. Nevertheless, the efficacy of these methods is often compromised by inaccurate attention mechanisms. As a result, the harmonization results tend to be interfered with by other irrelevant regions, thereby failing to produce visually pleasing results that effectively align with the related content in the reference.
As such, \emph{the second task of this work is to equip the harmonization method with a more accurate attention module}. 

Regarding the above two tasks, in this paper, we present Raiha, an innovative \textbf{r}etrieval-\textbf{a}ugmented framework for \textbf{i}mage \textbf{ha}rmonization.
Firstly, we propose a retrieval pipeline, making the retrieved reference have consistent illumination and similar objects with background region and foreground region, respectively.
For semantic comparison, the dense visual features from DVT~\cite{DVT} are considered, which helps find the objects similar to the foreground.
As for illumination, we turn to the retinex theory, which reveals that the appearance of an object can be decomposed as reflectance and illumination components.
In other words, any two terms of appearance, reflectance, and illumination can determine the other one.
Meanwhile, the reflectance is an inherent characteristic of the object.
In this way, finding the illumination-consistent reference becomes equivalent to matching the content and appearance.
Therefore, the DVT feature and HSV histogram are enough to determine the illumination consistency.
%
%

As for the attention mechanism, \mliu{we have visualized the attention maps of existing methods}, observing that the main issue is the influence of irrelevant regions.
To suppress the influence of such conditions, we propose a semantic-guided fusion module by reusing the DVT feature extracted during the image retrieval phase, which restricts the scope of attention to semantically similar regions.
It is worth noting that, besides the retrieved references, the target images can also act as ideal references for training Raiha in the proposed retrieval augmented setting, where random cropping and resampling operations are applied to avoid over-fitting problems.

With the proposed Raiha framework, superior image harmonization performance has been achieved under both non-reference and retrieval-augmented image harmonization settings.
A benchmark for retrieval-augmented image harmonization termed Raiharmony4 is also constructed by reorganizing the non-reference dataset iHarmony4.

%
The main contributions can be summarized as follows,
\begin{itemize}
    \item We propose a novel retrieval-augmented framework for image harmonization, Raiha, which provides reliable guidance by retrieving an external reference and concatenating it with the composite image.
    \item To make the guidance from retrieved reference reliable, we present a retrieval strategy for the image harmonization task. It guarantees the retrieved references not only have consistent illumination with the background but also contain content similar to the foreground.
    \item A semantic-guided fusion module and a data augmentation strategy are delicately designed to make Raiha effectively utilize the guidance information from retrieved references.
    \item Experiments on the image harmonization dataset demonstrate that our method can achieve state-of-the-art performance under both non-reference and retrieval-augmented settings.
\end{itemize}


\section{Related Works}
\subsection{Image Harmonization}

Traditional image harmonization methods usually extract hand-craft statistical features to make foreground objects have a visually consistent appearance with the background area. However, those methods have limited performance in complex scenarios.
DNN-based methods on image harmonization have made great progress in recent years. DoveNet~\cite{DoveNet} and BargainNet~\cite{Bargainnet} achieve domain consistency between foreground and background. Region-aware Adaptive Instance Normalization~\cite{RAIN} is proposed to align foreground features to normalized background features and achieve image harmonization. However, these methods usually predict dense pixel-to-pixel translation and thus cannot achieve high-resolution harmonization. 
Instead of predicting dense pixel-to-pixel translation~\cite{DoveNet,Bargainnet,RAIN}, some methods~\cite{DCCF,Harmonizer,CDTNet,SSATAAAI} are proposed to achieve image harmonization on high resolution. DCCF~\cite{DCCF} and Harmonizer~\cite{Harmonizer} predict comprehensible image filters, such as value, saturation, and hue. CDTNet~\cite{CDTNet} introduces RGB-to-RGB transformation while methods like S$^2$CRNet~\cite{SSATAAAI} predict piecewise curve mapping parameters.
Considering the limited scale of the existing dataset, Hang ~\textit{et al.} ~\cite{Scs-co} proposes to dynamically generate multiple negative samples under contrastive learning, which prevents from generating distorted harmonized results. Liu ~\textit{et al.}~\cite{LEMaRT} proposed to pretrain image harmonization network by leveraging large-scale unannotated image datasets with the LEMaRT pipeline. 

When the background contains similar content to that of the foreground, its appearance can provide reliable guidance for image harmonization. Based on this, Zhu \textit{et al.}~\cite{MRR} proposed the Localtion-to-Localtion module to fuse foreground location with the related background feature and the Patches-to-Location Module to further refine and adjust the foreground feature at the high-resolution level. Hang \textit{et al.}~\cite{Scs-co} design BAIN to obtain attention-weighted background feature distribution according to the foreground-background feature similarity. HDNet~\cite{HDNet} introduces Local Dynamic Module to find K local representations from background region according to foreground feature.

Nevertheless, attention-based methods are limited while the foreground-matched objects are unavailable in the background. Our method aims to retrieve external references as background extensions and further achieve visual harmony results. 

\subsection{Image Retrieval}
Existing image retrieval methods used off-the-shelf models or fine-tuning models for extracting distinguishable representation and then retrieving desired images for query images. 
Early methods~\cite{off-the-shelf,Neural_codes} use the fully connected layers of the pre-trained model to extract global features, whose descriptions cannot cover each object in an image. Multiple forward schemes are proposed to extract multiple object representations by adopting sliding window~\cite{ali2015baseline,do2017embedding}, spatial pyramid modeling~\cite{zhao2017spatial} and multiple regions~\cite{gordo2016deep} generated by networks~\cite{ren2015faster}. Some methods also use convolution layers~\cite{jimenez2017class} to produce spatial representation. 
Matching with global descriptors has high efficiency for feature extraction and similarity evaluation in image retrieval. Feature embedding methods like BoW~\cite{BoW}, VLAD~\cite{VLAD}, and FV~\cite{FV} are proposed to obtain aggregate extracted features into a global descriptor. 
However, global matching is not compatible with spatial correspondence retrieval~\cite{chen2022deep}. Therefore, local matching schemes~\cite{paulin2015local,tolias2016image} are proposed to evaluate the similarity across all local features. Considering local matching usually has intensive memory to store local features and low efficiency to summarize similarity, most image retrieval methods consist of initial filtering (global matching) and re-ranking (local matching) stages.

After constructing a dataset with well-defined labels for image retrieval, many methods~\cite{Neural_codes,cao2020unifying,lv2018retrieval,caron2018deep,revaud2019learning} are proposed to further improve distinguish performance of feature representation. The models after fine-tuned by Siamese loss~\cite{lin2017deephash,radenovic2018fine} and ranking loss~\cite{xiang2019multiple,min2020two} achieve better retrieval performance. These methods will be less infeasible while meeting the insufficient scale of the training set. Some unsupervised fine-tuning methods, mainly based on manifold learning~\cite{iscen2018mining,iscen2018fast} and clustering~\cite{tzelepi2018deep,caron2018deep}, are proposed to improve the ability to distinguish relevance representation between a query image and support gallery.

However, the above methods have not considered the illumination condition during image retrieval. Based on the off-the-shelf models, we propose an harmonization-oriented image retrieval method to retrieve consistent illumination from the support gallery.


\begin{figure*}[t]
    \centering
    \begin{minipage}{\linewidth}
        \begin{overpic}[width=.98\linewidth]{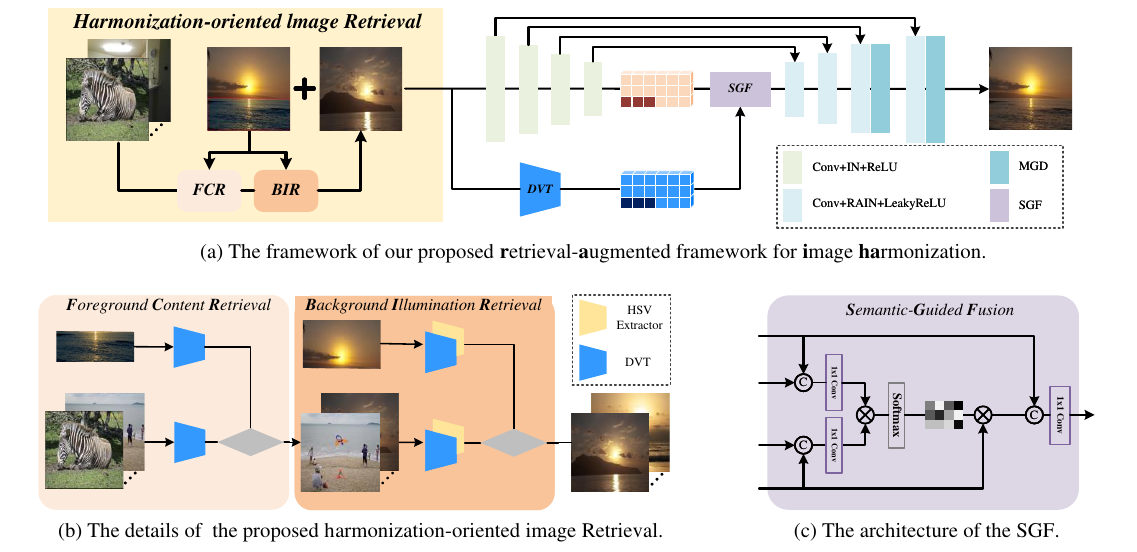}
				\put(8.0,34.5){$\mathcal{G}$}
                \put(23.0,37.5){\textcolor{white}{$\matx_f$}}
				\put(23.0,43.0){\textcolor{white}{$\matx_b$}}
				\put(19.8,35.5){$\tilde{\matx}$}
				\put(29.3,35.5){$\matx_r$}
				\put(53.5,28.8){[\textcolor[rgb]{0.0, 0.125, 0.375}{$\matc_f$},\textcolor[rgb]{0.20, 0.597, 1.0}{$\matc_b$},\textcolor[rgb]{0.20, 0.597, 1.0}{$\matc_r$}]}
				\put(90.0,44.5){$\hat{\matx}$}	
				\put(53.5,37.4){[\textcolor[rgb]{0.570 ,0.222, 0.191}{$\mate_f$},\textcolor[rgb]{0.98, 0.839, 0.73}{$\mate_b$},\textcolor[rgb]{0.98, 0.839, 0.73}{$\mate_r$}]}
				
				\put(60.5,5.0){[\textcolor[rgb]{0.98, 0.839, 0.73}{$\mate_b$},\textcolor[rgb]{0.98, 0.839, 0.73}{$\mate_r$}]}
				\put(60.5,8.8){[\textcolor[rgb]{0.20, 0.597, 1.0}{$\matc_b$},\textcolor[rgb]{0.20, 0.597, 1.0}{$\matc_r$}]}
				\put(64.0,14.8){\textcolor[rgb]{0.0, 0.125, 0.375}{$\matc_f$}}
				\put(64.0,18.7){\textcolor[rgb]{0.570 ,0.222, 0.191}{$\mate_f$}}
				\put(88.0,11.0){$\mate_w$}
				\put(96.5,11.8){$\mate_f'$}				
                
                \put(20.0,9.2){\scriptsize{$\mathit{Equ}$~\ref{equ:fcr}}}
                \put(43.0,9.2){\scriptsize{$\mathit{Equ}$~\ref{equ:bir}}}
				\put(7.0,15.5){$\matx_f$}
				\put(8.0,4.0){$\mathcal{G}$}	
				\put(29.0,4.0){$\mathcal{G}_{sc}$}
				\put(29.0,15.3){$\matx_{b}$}	
				\put(52.0,3.5){$\mathcal{G}_{sc\&ci}$}	
        
         \end{overpic}	

    \end{minipage}

      \caption{(a) presents the framework of our proposed Raiha, which solves the ill-posed issues when similar content is unavailable in the background $\matx_b$. Firstly, given a composite image $\tilde{\matx}$, we design a harmonization-oriented image retrieval method to produce proper a reference $\matx_r$ from the gallery $\mathcal{G}$. Specifically, a foreground content retrieval is designed to construct the gallery $\mathcal{G}_{sc}$, in which images contain similar content with foreground $\matx_f$. The background illumination retrieval is employed to construct the gallery $\mathbf{G}_{sc\&ci}$, in which images contain consistent illumination with the background $\matx_b$. The Semantic-Guided Fusion module (c) is also designed to enable Raiha to effectively utilize the reference. Finally, Raiha takes both $\tilde{\matx}$ and the reference $\matx_r$ as inputs to generate visually pleasant results $\hat{\matx}$.}
    \label{fig:Raiha framework}
\end{figure*}
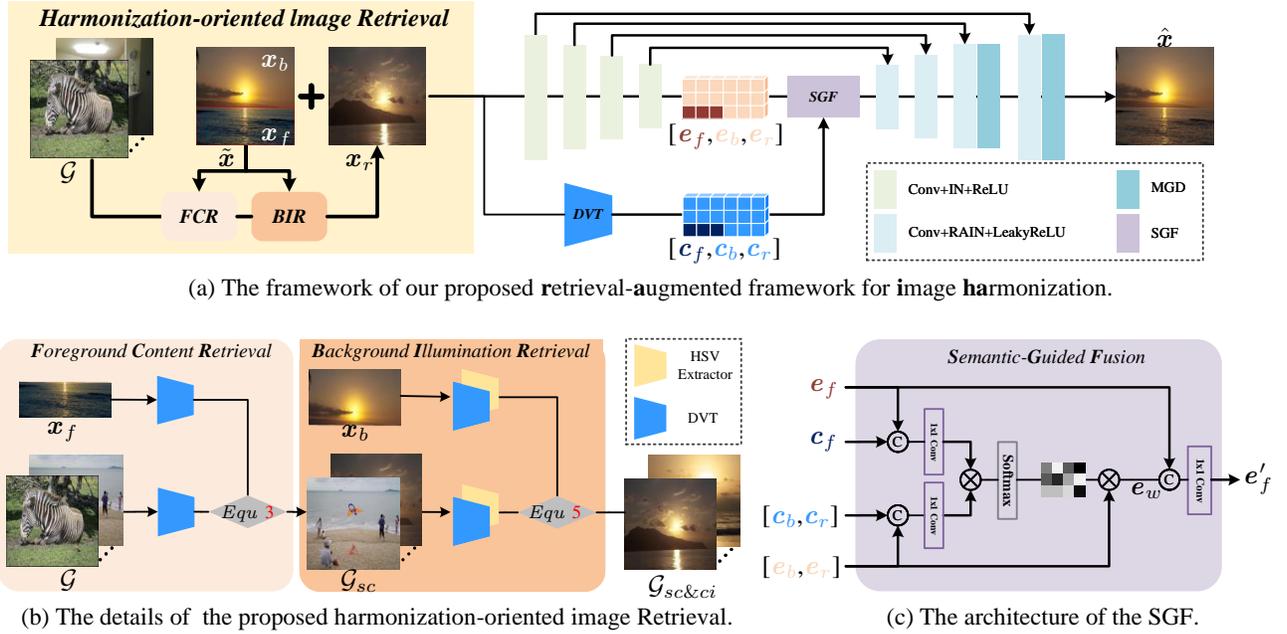

\section{Methodology}

\subsection{Overall Framework}\label{sec:method_overall}
Regarding the two main issues of existing image harmonization methods, \ie, the ill-posedness caused by the absence of reliable references and the unsatisfactory results due to inaccurate attention, we present a retrieval-augmented image harmonization (Raiha) framework, which is accordingly comprised of an image retrieval pipeline for obtaining suitable reference images and an image harmonization network equipped with a delicately designed semantic-guided fusion module.

As shown in \cref{fig:Raiha framework}, a composite image $\tilde{\matx}$ is made up of the background $\matx_b$ and an inserted foreground $\matx_f$.
The purpose of image harmonization is to obtain a harmonized image $\hat{\matx}$, where the appearance of the foreground is visually consistent with the background.
In our Raiha framework, we first look for suitable reference images (denoted by $\matx_r$) from a gallery $\mathcal{G}$.
Then, the final output $\hat{\matx}$ can be obtained by $\hat{\matx}=f(\tilde{\matx}, \tilde{\matm}, \matx_r; \theta_f)$, where $\tilde{\matm}$ denotes the mask indicating the foreground and background regions in $\tilde{\matx}$.
The details about the image retrieval pipeline and the image harmonization network are provided in \cref{sec:method_retrieval,sec:method_network}, respectively.

\begin{table*}[t]
    \centering
    \resizebox{\textwidth}{!}{
        \begin{tabular}{ccccccccccc}
            \toprule
            \multirow{2}{*}{Model} &  \multicolumn{2}{c}{HAdobe5k} & \multicolumn{2}{c}{HFlickr} & \multicolumn{2}{c}{HCOCO} & \multicolumn{2}{c}{Hday2night}& \multicolumn{2}{c}{Average} \\
            &MSE$\downarrow$&PSNR$\uparrow$&MSE$\downarrow$&PSNR$\uparrow$&MSE$\downarrow$&PSNR$\uparrow$&MSE$\downarrow$&PSNR$\uparrow$&MSE$\downarrow$&PSNR$\uparrow$\\
            \midrule
            Composite&345.54&28.16&264.35&28.32&69.37&33.94&109.65&34.01&172.47&31.63\\
            iS$^{2}$AM& 21.60&38.28&69.43&33.65&16.15 &39.40&40.39&37.87& 24.13&38.41\\
            RainNet&43.35&36.22 &110.59&31.64 &29.52&37.08&57.40&34.83&40.29& 36.12  \\
            BargainNet& 39.94 &35.34&97.32 & 31.34 & 24.84&37.03& 50.98 &35.67& 37.82 &35.88 \\
            Intrinsic&43.02&35.20&105.13&31.34&24.92& 37.16 &55.53& 35.96 &38.71&35.90 \\
            D-HT &38.53&36.88&74.51&33.13&16.89& 38.76&53.01&37.10&30.30&37.55\\
            Harmonizer&21.89&37.64&64.81&33.63&17.34& 38.77&33.14& 37.56&24.26&37.84 \\
            DCCF&19.90 & 38.27 &60.41 & 33.94&14.87 & 39.52&49.32 &37.88&22.05 & 38.50  \\
            CDTNet &20.62 &38.24&68.61&33.55 &16.25 & 39.15&36.72& 37.95 & 23.75 & 38.23\\
            PCT-Net& 20.86 & 40.11 & \textbf{{44.30}}& 35.13&\underline{{10.72}} &40.77 &44.74 & 37.65& 18.04 & 39.89   \\
            HDNet&\underline{{13.58}}&\underline{{41.17}} & 47.39 & \underline{{35.81}} & 11.60 & \underline{{41.04}} & \underline{{31.97}} & \underline{{38.85}} & \underline{{16.55}} & \underline{{40.46}} \\
            \midrule
            Raiha&\textbf{{13.32}} & \textbf{{41.55}}& \underline{{45.41}}&\textbf{{36.44}} & \textbf{{10.53}}&  \textbf{{41.82}}&\textbf{{30.38}} & \textbf{{39.77}}&   \textbf{{15.60}} &  \textbf{{41.10}}\\
            \bottomrule
    \end{tabular}}
    \caption{Quantitative comparison across four sub-datasets of iHarmony4~\cite{DoveNet}. The top two performances are shown in \textbf{{bold}} and \underline{underline}. $\uparrow$ means the higher the better, and $\downarrow$ means the lower the better. The Raiha results are reported under the non-reference setting.}
    \label{tab:iHarmomny4}
\end{table*}

\subsection{Harmonization-oriented Image Retrieval}\label{sec:method_retrieval}
Existing image retrieval methods typically focus on semantic similarities. However, since image harmonization tasks modifies mainly the appearance of the image, merely considering image content cannot guarantee the usability of retrieval results.
Generally speaking, defining the content and illumination of an image by $\matc$ and $\mati$, respectively, a desirable reference image should have the following properties.
\begin{equation}
	\matc_r \approx \matc_f \quad \mathrm{and} \quad \mati_r \approx \mati_b,
\end{equation}
where the subscripts $f$, $b$, and $r$ denotes foreground, background, and reference, respectively.

\paragraph{Foreground Content Retrieval.}

Thanks to the development of self-supervised learning techniques~\cite{CLIP,DINOV2,DVT}, we can extract dense semantic features of an image via pre-trained models like DVT~\cite{DVT} (denoted by $f_\ii{DVT}$), \ie,
\begin{equation}
	\matc_f = f_\ii{DVT}(\matx_f), \matc_r = f_\ii{DVT}(\matx_r),
\end{equation}
where $\matc_f=\{\matc_f^i\}_{i=1}^M$, $\matc_r=\{\matc_r^j\}_{j=1}^N$ due to the patchify operation in practice.
Then, we consider two images to contain semantically similar content, once we have
\begin{equation}
	\langle \matc_f^i, \matc_r^j \rangle \ge \epsilon_{c},
 \label{equ:fcr}
\end{equation}
for any $i$ and $j$, where $\langle\cdot,\cdot\rangle$ denotes cosine similarity, and $\epsilon_c$ is a threshold empirically set to 0.7 in this paper.

\paragraph{Background Illumination Retrieval.}
In terms of consistent illumination retrieval, we refer to the Retinex theory, which explains that the appearance of an object can be decomposed as reflectance and illumination components. While knowing any two of these components, the remain one can be determined.
Thus, we extract the content $\matc$ and appearance $\mata$ of the background and the reference, \ie,
\begin{equation}
    \matc_b = f_\ii{DVT}(\matx_b), \mata_b = f_\ii{HSV}(\matx_b), \mata_r = f_\ii{HSV}(\matx_r),
\end{equation}
and we consider the reference images with the following properties to be consistent with the background image illumination, \ie,
\begin{equation}
	\langle \matc_b^i, \matc_r^j \rangle \ge \epsilon_{c}, \langle \mata_b^i, \mata_r^j \rangle \ge \epsilon_{a},
\label{equ:bir}
\end{equation}
where $\mata_b^i \in \mata_b$ and $\mata_r^j \in \mata_r$ denote the HSV histogram of two patches in background $\matx_b$ and reference $\matx_r$, respectively. $\epsilon_a$ is a threshold empirically set to 0.9 in this paper.

\subsection{Model Design}\label{sec:method_network}
\paragraph{Network Architecture.}
Existing image harmonization methods typically deploy a U-Net~\cite{unet} structure, where the foreground and background features interact through cross attention mechanisms, thereby adjusting the foreground features based on the background image.
The calculation of the attention map can be formulated by,
\begin{equation}
    \bb{A} = \phi(\mate_f) \times \phi(\mate_b),
    \label{eqn:attn_vanilla}
\end{equation}
where $\phi$ denotes linear (or $1\times1$ convolution) layers, $\mate$ means the features extracted by the encoder, and $\times$ means matrix multiplication.
In our Raiha framework, we follow the overall scheme of existing methods and extend them from two aspects, \ie, introducing the retrieved reference image $\matx_r$ and considering more accurate attention.
The purpose of introducing $\matx_r$ is to compensate for the content absence of the background regions. Therefore, in our harmonization network, we regard $\matx_r$ as a supplement of $\matx_b$ and treat them equally.
Therefore, the feature of $\matx_r$ is also introduced in \cref{eqn:attn_vanilla}, \ie,
\begin{equation}
    \bb{A} = \phi(\mate_f) \times \phi([\mate_b, \mate_r]_s),
    \label{eqn:attn_ext}
\end{equation}
where $[\cdot, \cdot]_s$ denotes the concatenation operation on the spatial dimension.
As for the attention, we introduce the semantic information as guidance, and a semantic-guided fusion module is presented for effective feature interaction.

\paragraph{Semantic-Guided Fusion Module.}

Due to the lack of understanding of image contents, existing methods often generate inaccurate attention, and the harmonized results are easily interfered with by other irrelevant areas.
To alleviate this issue, we design a \mliu{\textbf{S}emantic-\textbf{G}uided \textbf{F}usion module} to encourage Raiha to effectively adjust foreground region appearance according to the related region in the retrieved reference.
Specifically, we reuse the dense visual features $(\matc_f,\matc_b,\matc_r)$ during the harmonization-oriented image retrieval pipeline, which contain the content information of the images extracted by DVT~\cite{DVT} and can serve as semantic guidance for the attention map calculation, \ie,
\begin{equation}
    \bb{A} = \phi([\mate_f, \matc_f]_c) \times \phi([[\mate_b, \mate_r]_s, [\matc_b, \matc_r]_s]_c),
    \label{eqn:attn_ours}
\end{equation}
where $[\cdot, \cdot]_c$ denotes the concatenation operation on the channel dimension.
In this way, the attention module can rely on semantic guidance to avoid incorrect attention, and the weighted features can be represented by,
\begin{equation}
    \mate_w = \sigma(\bb{A})\times[\mate_b, \mate_r]_s,
\end{equation}%
where $\sigma$ denotes the softmax operation, making Raiha pay more attention to the foreground-similar content in both background and reference regions, instead of harmonization-unrelated information.
Finally, the foreground features can be modulated via,
\begin{equation}
    \mate_f' = \phi([\mate_f, \mate_w]_c)
\end{equation}

\paragraph{Data Augmentation.}

By employing the harmonization-oriented image retrieval method, multiple retrieved references can be generated for each composite image. 
Besides these retrieved references, we utilize the target image as a reference for each composite image. Target images inherently have consistent illumination with composite images. However, directly using the target image to provide supervision information leads to overfitting problems. To mitigate this, we apply random cropping, flipping, and resizing operations to the target image to create diverse references for the composite images. Note that augmented references encompass the foreground region.
Augmented references are designed to enhance Raiha's ability to efficiently utilize the guidance from references, enabling it to adjust the appearance of the foreground accordingly. However, training Raiha only with these augmented references may lead to the model completely relying on the information in the reference image, thereby altering the intrinsic appearance information of the foreground, such as color. Thus the retrieved references and augmented references should be used simultaneously to ensure the Raiha solves the ill-posedness issue and achieves visually pleasant results under the retrieval-augmented setting.

\paragraph{Learning Objective.}
Using the above references, Raiha can be trained easily and then achieve harmonization under the guidance of the retrieved references. Following~\cite{HDNet}, we only employ the foreground MSE loss as our loss function:
\begin{equation}
    \mathcal{L} = \frac{\left \| \hat{\matx} - \matx \right \|_2^2}{\max\{ \epsilon_f, {\sum}{\matm_f} \}}
\end{equation}
where $\matx$ denotes the target image, $\sum\matm_f$ means the area of the foreground regions, $\epsilon_f$ is a small value to avoid numeric overflow.
Considering that there may be no proper references for some cases, we also consider scenarios with no reference images available.
During the training phase, non-reference and retrieval-augmented settings are randomly sampled for \mliu{each training iteration}.

\begin{table*}[t]
    \centering
    \resizebox{\textwidth}{!}{
        \begin{tabular}{ccccccccccc}
            \toprule
            \multirow{2}{*}{Model} &  \multicolumn{2}{c}{RAHAdobe5k} & \multicolumn{2}{c}{RAHFlickr} & \multicolumn{2}{c}{RAHCOCO} & \multicolumn{2}{c}{RAHday2night}& \multicolumn{2}{c}{Average} \\
            &MSE$\downarrow$&PSNR$\uparrow$&MSE$\downarrow$&PSNR$\uparrow$&MSE$\downarrow$&PSNR$\uparrow$&MSE$\downarrow$&PSNR$\uparrow$&MSE$\downarrow$&PSNR$\uparrow$\\
            \midrule
            Composite&388.59 &26.74 & 338.06& 25.24&89.20 & 31.71& 95.36 & 29.27 & 186.57 & 29.89\\
            iS$^{2}$AM& 22.44 & 37.79 & 76.22  &31.47 & 16.6 &  38.45 & 41.29 & 33.37 & 22.99 & 37.69 \\
            RainNet & 43.92 & 34.83& 167.53  &  28.29& 38.81 & 34.82 & 62.47 & 32.07 &50.34  & 34.29\\ 
            BargainNet& 41.29 & 33.88 & 127.42 & 28.80 & 30.59 &35.59 & 56.46  &  30.97 & 41.14 & 34.57 \\ 
            Intrinsic& 34.00   & 35.44 & 124.61 & 29.21 & 26.6  & 36.31 & 56.05 & 31.59 & 36.42 & 35.49 \\
            D-HT& 38.25 &  35.46 & 84.18 &  30.98 & 19.42 &  37.74 & 40.96 &33.40 & 29.56 & 36.58 \\
            Harmonizer&  29.74 &  36.74 &  85.16 &  30.88 & 21.82 &  37.34  & 43.72 &  32.37  & 29.01 & 36.63 \\ 
            DCCF     & 21.48 &  36.77  & 88.31 & 30.91 &18.25 & 38.01& 41.35 & 33.45 & 24.75 &  37.09 \\
            CDTNet &21.75 & 36.45 &96.00  & 30.56 &19.75 &37.55  &36.92 & 33.46 & 26.35& 36.68\\
            PCT-Net & 20.51 &  39.12 & \underline{{61.17}} & 32.30  & \underline{{12.89}} &  39.34 & 33.30 & 34.21  & 18.83  & 38.68 \\
            HDNet& \textbf{{13.27}}& \underline{{39.42}} & {71.72} & \underline{{32.46}} &{{14.42}} & \underline{{39.36}} & \underline{{30.70}} & \underline{{34.68}} & \underline{{18.72}} & \underline{{38.80}}   \\
            \midrule
            Raiha & 
            \makecell{\underline{{13.35}}\\ \stdvu{$\pm$0.21}}&
            \makecell{\textbf{{40.26}}\\~\stdvu{$\pm$0.04}}&
            \makecell{\textbf{{51.54}}\\~\stdvu{$\pm$0.57}}&
            \makecell{\textbf{{33.47}}\\~\stdvu{$\pm$ 0.04}}&
            \makecell{\textbf{{12.40}}\\~\stdvu{$\pm$ 0.02}}&
            \makecell{\textbf{{40.01}}\\~\stdvu{$\pm$ 0.02}}&
            \makecell{\textbf{{30.16}}\\~\stdvu{$\pm$ 0.58}}&
            \makecell{\textbf{{35.47}}\\~\stdvu{$\pm$ 0.09}}&
            \makecell{\textbf{{15.86}}\\~\stdvu{$\pm$ 0.01}}&
            \makecell{\textbf{{39.58}}\\~\stdvu{$\pm$ 0.02}}\\
				\bottomrule
		\end{tabular}}
		\caption{Quantitative comparison across four sub-datasets of RAHarmony4 under the setting of retrieval-augmented. The top two performances are shown in \textbf{bold} and \underline{{underline}}. $\uparrow$ means the higher the better, and $\downarrow$ means the lower the better. The Raiha results are reported as the average of 5 runs.}
		\label{tab:RAHarmomny4} 

\end{table*}
\begin{figure*}[t]
    \centering
    \begin{minipage}{0.11\linewidth}
        \includegraphics[width=\linewidth]{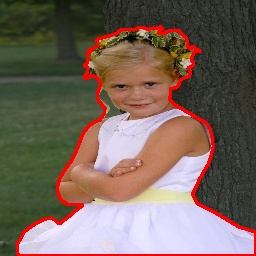}\\
        \includegraphics[width=\linewidth]{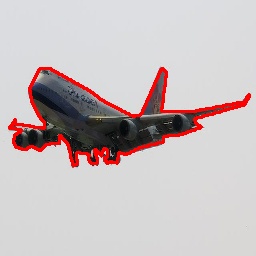}\\
        \includegraphics[width=\linewidth]{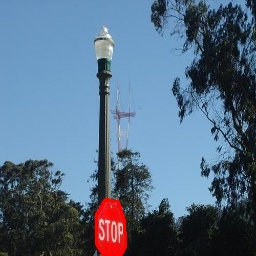}
        \caption*{ Input}
    \end{minipage}
    \begin{minipage}{0.11\linewidth}
        \includegraphics[width=\linewidth]{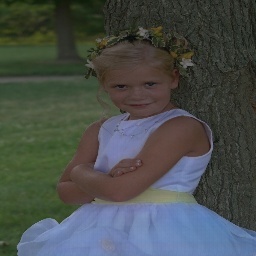}\\
        \includegraphics[width=\linewidth]{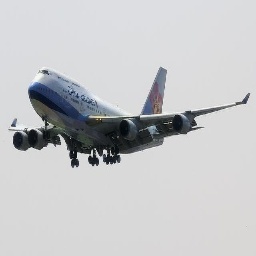}\\
        \includegraphics[width=\linewidth]{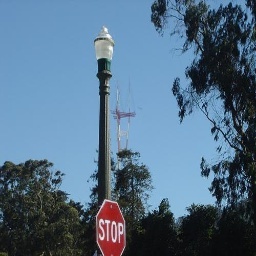}
        \caption*{ DCCF}
    \end{minipage}
    \begin{minipage}{0.11\linewidth}
        \includegraphics[width=\linewidth]{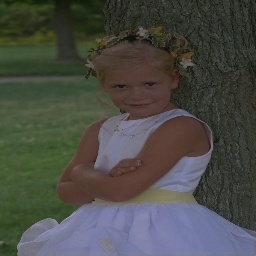}\\
        \includegraphics[width=\linewidth]{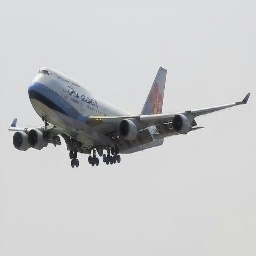}\\
        \includegraphics[width=\linewidth]{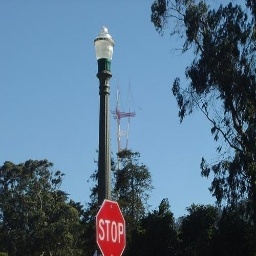}
        \caption*{ CDTNet}
    \end{minipage}
    \begin{minipage}{0.11\linewidth}
        \includegraphics[width=\linewidth]{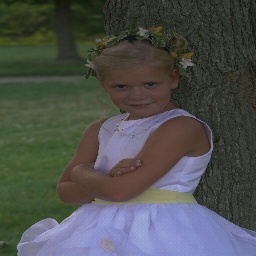}\\
        \includegraphics[width=\linewidth]{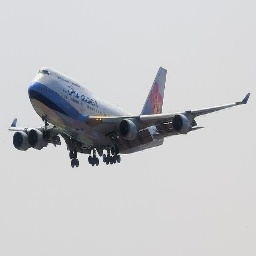}\\
        \includegraphics[width=\linewidth]{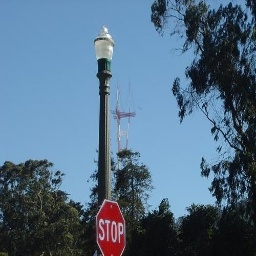}
        \caption*{ PCTNet}
    \end{minipage}
    \begin{minipage}{0.11\linewidth}
        \includegraphics[width=\linewidth]{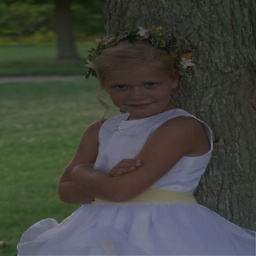}\\
        \includegraphics[width=\linewidth]{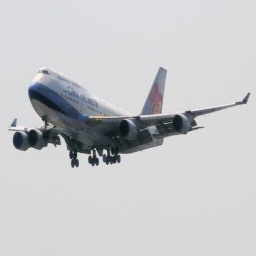}\\
        \includegraphics[width=\linewidth]{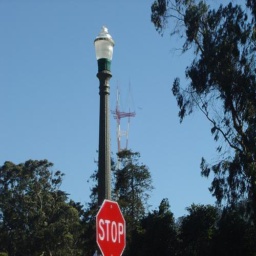}\
        \caption*{ HDNet}
    \end{minipage}
    \begin{minipage}{0.11\linewidth}
        \includegraphics[width=\linewidth]{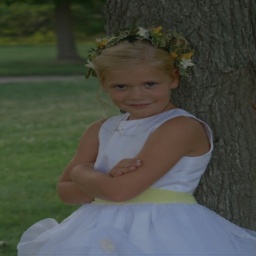}\\
        \includegraphics[width=\linewidth]{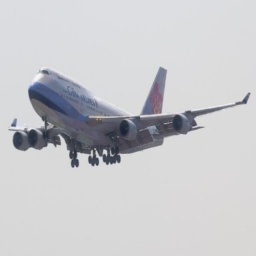}\\
        \includegraphics[width=\linewidth]{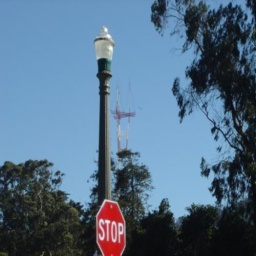}
        \caption*{ Raiha}
    \end{minipage}
    \begin{minipage}{0.11\linewidth}
        \includegraphics[width=\linewidth]{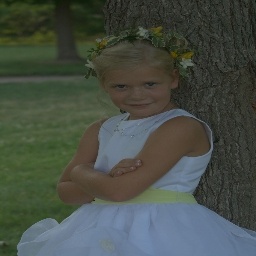}\\
        \includegraphics[width=\linewidth]{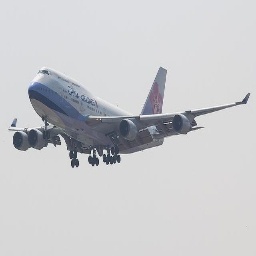}\\
        \includegraphics[width=\linewidth]{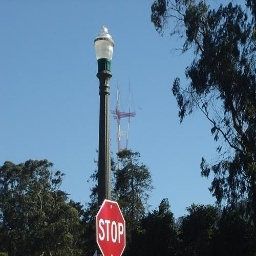}
        \caption*{ Target}
    \end{minipage}
    \begin{minipage}{0.11\linewidth}
        \includegraphics[width=\linewidth]{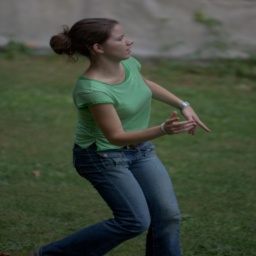}\\
        \includegraphics[width=\linewidth]{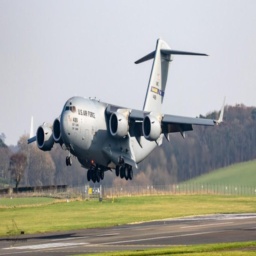}\\
        \includegraphics[width=\linewidth]{sotas_results/c23542_1809288_1_ours_ref.jpg}
        \caption*{ Reference}
    \end{minipage}
    \caption{Qualitative comparison results on the RAHarmony4 dataset (the retrieval-augmented setting of the iHarmony dataset). Raiha produces more visually consistent results than other SOTA methods.}
    \label{fig:comparison iHarmony4}
\end{figure*}

\section{Experiments}

\subsection{Datasets} 
We conduct experiments on the iHarmony4 dataset~\cite{DoveNet} by following the setting as Bargainnet~\cite{Bargainnet} and RAIN~\cite{Bargainnet,RAIN}. iHarmony4 is composed of 73,146 image pairs which contain four subsets: HAdobe5k, HFlickr, HCOCO, and Hday2night. Each sample in iHarmony4 consists of a target image, a foreground mask, and a composite image. We follow the same split settings in DoveNet~\cite{DoveNet}, where 65,742 images are for training and the remaining 7,404 images are for testing. The iHarmony4 dataset is utilized to evaluate the performance of Raiha under the non-reference setting. For a fair comparison with other methods and to investigate the effects of the retrieval-augmented setting, we collect target images from the training set of iHarmony4 to construct the RAHarmony4 dataset, employing the harmonization-oriented retrieval pipeline. The RAHarmony4 testing set consists of 2,248 images.

\subsection{Implementation and Evaluation Details}

Raiha is optimized by Adam algorithm by Adam optimizer with $\beta_1=0.9$, and $\beta_2=0.999$. We train Raiha for 60 epochs with a batch size of 48, and the initial learning rate is set to 0.004. The linear learning rate decay strategy is also employed after 40 epochs, with the final learning rate set to 0. All images are resized to $256\times256$. We use PyTorch~\cite{paszke2019pytorch} to implement our models with two A6000 GPUs. 
We adopt MSE and Peak Signal-to-Noise Ratio (PSNR) to evaluate the quantitative performance on both the iHarmony4 and RAHarnomy4 datasets. While a composite image has multiple references produced by our retrieval method, we randomly select a reference and calculate the metrics. The results are reported as the average of 5 runs.

\subsection{Comparison with The Other Methods}
We compare our method with state-of-the-art methods, including iS$^{2}$AM~\cite{sofiiuk2021foreground}, RainNet~\cite{RAIN}, Bargainnet~\cite{Bargainnet}, Intrinsic~\cite{GuoZJGZ21}, D-HT~\cite{IHT}, CDTNet~\cite{CDTNet}, Harmonizer~\cite{Harmonizer}, DCCF~\cite{DCCF}, PCTNet\cite{PCTNet} and HDNet~\cite{liu2023lemart}. Note that all methods are trained and tested in the resolution of $256\times256$. 

Firstly, we conduct comparison experiments on the iHarmony4 dataset, where Raiha only takes composite images as inputs. As shown in~\cref{tab:iHarmomny4}, our proposed Raiha can achieve superior performance against the state-of-the-art methods under the non-reference setting. We further evaluate the performance of the RAHarnomy4 dataset. Note that comparison methods only take composite images as inputs. As shown in~\cref{fig:comparison iHarmony4}, the retrieved references contain similar content with foreground and consistent illumination with the background. Raiha adjusts foreground appearance according to its related locations in the retrieved reference, and further, Raiha achieves improvement on the RAHarmony4 dataset compared to other methods (shown in~\cref{tab:RAHarmomny4}).

\subsection{Ablation Study}

\noindent\textbf{Effects of different training set.}
We train Raiha with references that are generated by the proposed retrieval method and augmentation separately. Without using retrieved references to train the Raiha (denoted as "Raiha trained w/o Retrieval"), Raiha achieves less performance. The reason is that Raiha blindly relies on the information from references, and tends to change the foreground's intrinsic appearance. In other words, Raiha is less robust and cannot distinguish whether guidance from the retrieved reference is reliable for adjusting the foreground appearance. 
In the meantime, without using reference augmented from target images (denoted as "Raiha trained w/o Augmentation"), the performance of Raiha drops. The reason is that the provided guidance from the reference in the training phase often contains partial misalignment with the supervision from target images. In this case, Raiha has less ability to utilize the reference effectively.

\noindent\textbf{Effects of the SGF module.}
Without the SGF module, although similar content is available, the harmonized result is easily interfered with by irrelevant content in the reference. Quantitative results are shown in~\cref{tab:ablation}.

\noindent\textbf{Effects of different retrieved references.}
As shown in~\cref{tab:ablation}, without using retrieved reference, the performances of Raiha drop on the RAHarmony4 dataset. The quantitative results, $\stdvu{\pm0.01}$ and $\stdvu{\pm0.02}$ in terms of MSE and PSNR, also indicate that the overall quality of multiple references generated by our proposed retrieval method is reliable for image harmonization. Besides the proposed retrieval method, this stable improvement is also contributed by the robust utilization of reference images. While the foreground-similar content in reference has a different color appearance from the foreground, the harmonized result still keeps the intrinsic color of the foreground. 
While the range of $\epsilon_{a}$ decreases, the foreground-similar content in inconsistent illumination references provides less reliable guidance. The performances are also slightly lower than the non-reference setting on the RAHarmony dataset. It proves that illumination consistency is an essential criterion in the retrieval process. 
The qualitative results of Raiha and its variants are presented in the supplementary material.
\begin{table}[t]
\small
\renewcommand\arraystretch{1.2}
\label{tab:ablation}
\centering
\resizebox{0.45\textwidth}{!}{
    \begin{tabular}{cccc}
        \toprule
        \multicolumn{2}{c}{\multirow{2}{*}{Method}} &  \multicolumn{2}{c}{RAHarmony4} \\
        & &MSE$\downarrow$&PSNR$\uparrow$\\	
        \midrule
        \multicolumn{2}{c}{Raiha trained w/o Augmentation}  &   16.91~\stdvu{$\pm$ 0.06}   &   39.33~\stdvu{$\pm$ 0.02}  \\ 
        \midrule
        \multicolumn{2}{c}{Raiha trained w/o Retrieval} &     17.61~\stdvu{$\pm$ 0.06}   &  39.21~\stdvu{$\pm$ 0.02} \\
        \midrule
        \multicolumn{2}{c}{Raiha w/o SGF}   &  15.99~\stdvu{$\pm$ 0.09} & 39.51~\stdvu{$\pm$ 0.01} \\	
        \midrule
        \multirow{2}{*}{\makecell{Raiha w/ reference \\ (inconsistent illumination) }} & $0.7\le \epsilon_a<0.8$ & 16.39~\stdvu{$\pm$ 0.09} &  39.43~\stdvu{$\pm$ 0.01}  \\
        &$0.8\le \epsilon_a<0.9$&  16.31~\stdvu{$\pm$ 0.12} &   39.46~\stdvu{$\pm$ 0.00}   \\	
        \midrule
        \multicolumn{2}{c}{Raiha non-reference}&  16.34 & 39.47  \\	
        \midrule
        \multicolumn{2}{c}{Raiha}             &  15.86~\stdvu{$\pm$ 0.01}&  39.58~\stdvu{$\pm$ 0.02}	\\
        \bottomrule
\end{tabular}}
\caption{Quantitative results of Raiha and its variants on RAHarmony4 dataset (retrieval-augmented setting from iHarmony4).}
\label{tab:ablation}
\end{table}

\section{Conclusion}
In this paper, we propose an innovative retrieval-augmented method for image harmonization, Raiha, which retrieves proper reference and guides foreground appearance adjustment while foreground-similar content is not available in the original background image. 
To guarantee the retrieved reference provides reliable guidance, we design an efficient harmonization-based image retrieval method. Specifically, we extract dense visual features via a pre-trained model and HSV histogram features. Firstly, we measure the content similarity between the foreground area and support images in the gallery, to make the retrieved images contain content similar to that of the foreground. Additionally, we verify the illumination consistency by examining the existence of two objects that both have similar content and consistent appearance within foreground and support images, respectively. 
To alleviate the interference of irrelevant information in the reference, we designed a Semantic-Guided Fusion module, which utilizes the content prior to calculating the similarity map. We also employ a data augmentation strategy by leveraging the existing dataset, to make Raiha effectively utilize the reference by providing consistent supervision with the target image.
Our proposed Raiha can outperform the state-of-the-art methods in both non-reference and retrieval-augmented settings of image harmonization tasks.

{\small
	\bibliographystyle{IEEEtran}
	\bibliography{egbib}
}


%
%
%
%

\vfill

\end{document}